\documentclass[11pt,a4paper]{article}
\usepackage{amsmath}
\usepackage[utf8]{inputenc}
\usepackage[hyperref]{emnlp2018}
\usepackage{times}
\usepackage{latexsym}

\usepackage{url}
\usepackage{graphicx}
\usepackage{mathtools}

\aclfinalcopy 

\title{Gated Recurrent Neural Network Approach for Multilabel Emotion Detection in Microblogs}

\author{Prabod Rathnayaka$^1$, Supun Abeysinghe$^1$, 
        Chamod Samarajeewa$^1$,\\ 
    {\bf Isura Manchanayake$^1$, Malaka J.Walpola$^1$, Rashmika Nawaratne$^2$, } \\ 
    {\bf Tharindu Bandaragoda$^2$ \and Damminda Alahakoon$^2$}  \\
    $^1$Department of Computer Science and Engineering,\\ 
    University of Moratuwa, Sri Lanka.\\
    $^2$Research Centre for Data Analytics and Cognition,\\ 
    La Trobe University Victoria, Australia.\\
  {\tt $^1$\{prabod.14,supun.14,chamod.14,isura.14,malaka\}@cse.mrt.ac.lk}\\
  {\tt $^2$\{b.nawaratne,t.bandaragoda,d.alahakoon\}@latrobe.edu.au }\\
  }

\begin{document}
\maketitle
\begin{abstract}
People express their opinions and emotions freely in social media posts and online reviews that contain valuable feedback for multiple stakeholders such as businesses and political campaigns. Manually extracting opinions and emotions from large volumes of such posts is an impossible task. Therefore, automated processing of these posts to extract opinions and emotions is an important research problem. However, human emotion detection is a challenging task due to the complexity and nuanced nature. To overcome these barriers, researchers have extensively used techniques such as deep learning, distant supervision, and transfer learning. In this paper, we propose a novel Pyramid Attention Network (PAN) based model for emotion detection in microblogs. The main advantage of our approach is that PAN has the capability to evaluate sentences in different perspectives to capture multiple emotions existing in a single text. The proposed model was evaluated on a recently released dataset and the results achieved the state-of-the-art accuracy of 58.9\%.
\end{abstract}

\section{Introduction}
Emotions are an integral part of human life which in turn affects human decision-making process and human thinking patterns. People often tend to express their opinions freely in social media compared to other means. Typically microblogs written by social media users tend to have effects from their emotions towards the topic they discuss or the emotions they feel at the moment. Hence, the linguistic features of these texts are highly dependent on the emotion and therefore can be used to extract the underlying emotion. Identifying underlying emotions of microblogs is useful in understanding author's opinions. This becomes beneficial in natural language applications in diverse fields including marketing, political campaigns, governing and human behavioral analysis.

Sentiment analysis can be considered as a fundamental type of emotion analysis. Though there have been a significant volume of research about sentiment analysis over the years, research about emotion detection and analysis has not gained much attention. A potential reason for that is complex and subtle behavior of human emotions compared to simple negative/positive sentiments. Research literature shows that there have been attempts to tackle this challenge using distant supervision techniques where emojis or hashtags present in the text are considered as indicators of emotions. However, such data can be noisy and somewhat unreliable, which affects the accuracy of such approaches.

Recently \newcite{mohammad2018semeval} has released a significantly large dataset as a SemEval 2018 task. Recent advancements in deep learning for natural language processing shows, given enough data, it is often possible to develop a model which achieves a reasonable level of accuracy. Frequently, attention mechanisms are employed in such deep learning models. In this paper, we propose a novel attention mechanism \emph{Pyramid Attention Network}(PAN) which has the ability to attend sentences from different perspectives. This becomes vital in emotion detection since a single micro-blog can contain multiple emotions and it needs to be considered in different perspectives to extract all the inherent emotions. State-of-the-art results achieved in our experiments is a good indication of the effectiveness of this approach.

\section{Related Work}
\newcite{mohammad2012emotional}, \newcite{mohammad2015using}, \newcite{wang2012harnessing}, \newcite{volkova2016inferring} and \newcite{abdul2017emonet} used distant supervision learning approach to acquire Twitter data with emotion related hashtags and emoticons. \newcite{abdul2017emonet} presents the largest dataset out of above all and they conducted a validation of the collected dataset using human annotators. A sample of 5600 tweets from the dataset has been checked and the study revealed only 61.37\% of the inferred emotions were relevant. Hence model evaluations for emotion detection done using such noisy datasets can be inaccurate.

Another approach for emotion detection is to use transfer learning. The frequent usage of emojis can be seen in social media posts to express the emotion associated with the text. DeepMoji \cite{felbo2017using} has addressed emotion prediction using a proposed variant of transfer learning called ‘chain-thaw’ where they used a pre-trained model which predicts emoji occurrences in microblogs. The model is based on a bidirectional Long Short Term Memory (LSTM) Network \cite{hochreiter1997long} which was trained using a huge dataset of 1 billion tweets.

LSTMs \cite{hochreiter1997long} and Gated Recurrent Units (GRU) \cite{cho2014learning, chung2015gated} which are variants of Recurrent Neural Networks (RNN), and Convolutional Neural Networks (CNN) have shown state-of-the-art results in text classification tasks including sentiment analysis \cite{ren2016context, liu2015multi, tai2015improved, tang2015document, zhang2016gated, kalchbrenner2014convolutional, kim2014convolutional, zhang2015character}.
\newcite{baziotis2018ntua} got top results in the SemEval task using transfer learning approach combined with Deep Attentive RNNs \cite{bahdanau2014neural}. Among other top results, \newcite{park2018plusemo2vec} used a transfer learning based approach whereas \newcite{kim2018attnconvnet} and \newcite{rozental2018amobee} used attention based approaches.

\section{Methodology}
First, the raw tweets are preprocessed, then mapped into a continuous vector space using an embedding layer. Then two stacked layers of Bidirectional GRUs \cite{cho2014learning} are used for feature extraction. This is followed by the novel attention mechanism where weighted average of the extracted features are taken. Finally, a dense layer maps this to eleven emotion categories using a sigmoid activation function.

\subsection{Preprocessing}
Preprocessing is a key step in the model as it affects the accuracy of the model significantly. We have used ekphrasis tool introduced by \newcite{baziotis-pelekis-doulkeridis:2017:SemEval2} for tweet preprocessing. Tweet tokenizing, word normalization, spell correcting and word segmentation for hashtags are done as preprocessing steps.

\begin{figure}[h]
    \centering
    \includegraphics[width=0.5\textwidth]{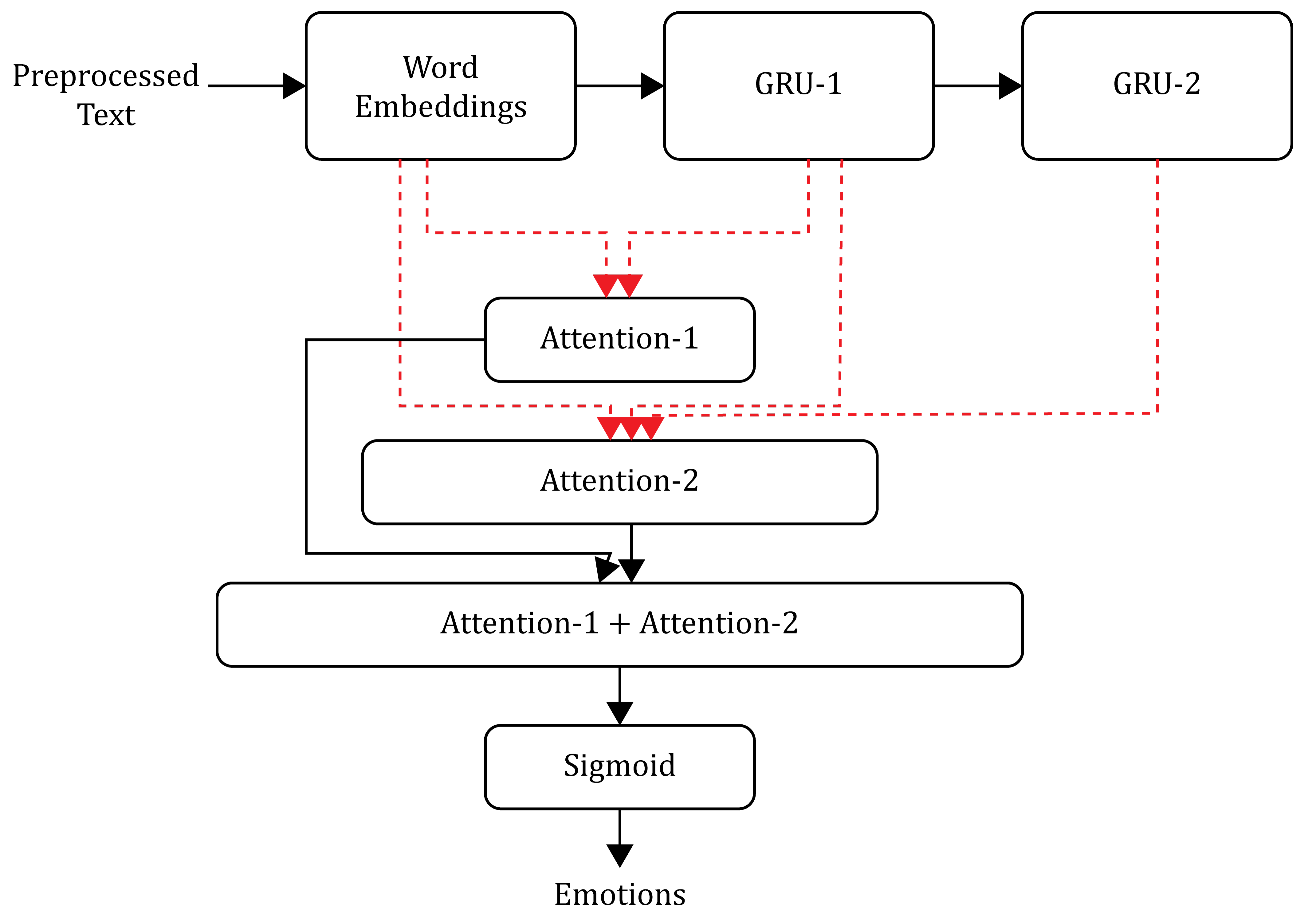}
    \caption{Overall model architecture}
    \label{fig:model}
\end{figure}

\subsection{Embedding Layer}
Given a sentence $S = [s_1,s_2,.,s_t,..,s_T]$ where $s_t$ is the one hot vector representation of word at position $t$, the words are embedded into a continuous vector space using an embedding matrix $W_e$.
\begin{align}
X = SW_e
\end{align}
$X$ serves as the embedding of the input sequence and is fed to the first GRU Layer. We used pre-trained Glove \cite {pennington2014glove} word embeddings (300 dimension) in our model. These pre-trained word vectors are kept frozen so that they are not updated during back-propagation. Using pre-trained word embeddings improved the model performance notably.

\subsection{Bidirectional GRU Layers}
Gated Recurrent Unit (GRU) \cite{cho2014learning} is an improved version of a standard Recurrent Neural Network designed to overcome the vanishing gradient problem and exploding problem \cite{bengio1994learning,hochreiter1997long}. GRU uses two special gating mechanisms called, update gate ($z_t$) and reset gate ($r_t$) to decide which information is passed to the output. Update gate helps the model to determine how much of past information is passed on to the future steps. Reset gate helps the model to determine how much of past information to forget. Then the output of position $t$ is created using concatenating forward and backward hidden states $(\overrightarrow{h_t} ,  \overleftarrow{h_t})$. Hidden size (output size) of both GRU layers is set to 50. $W$ and $b$ in (2), (3) and (4) corresponds to the weights and biases for the gates.
\begin{gather}
\begin{align}
r_t &= \sigma(W_{ir} x_t + b_{ir} + W_{hr} h_{(t-1)} + b_{hr}) \\
z_t &= \sigma(W_{iz} x_t + b_{iz} + W_{hz} h_{(t-1)} + b_{hz}) \\
n_t &= \tanh(W_{in} x_t + b_{in} + r_t (W_{hn} h_{(t-1)}+ b_{hn})) \\
h_t &= (1 - z_t) n_t + z_t h_{(t-1)}
\end{align}
\end{gather}

\subsection{Pyramid Attention Network}
Neural Networks with Attention has shown success in a variety of tasks such as machine translation \cite{luong2015effective}, question answering \cite{yang2016stacked}, image captioning \cite{you2016image} and sentiment analysis \cite{yang2016hierarchical}.
Aforementioned tasks such as sentiment analysis which can infer the sentiment based on a set of keywords, can directly use attention to focus on a specific part of a sentence and decide the sentiment. However, in multi-label emotion classification, attending to a part of a sentence can infer some emotions but could not capture every emotion embedded in the sentence. Following is a tweet extracted from the dataset.

\emph{The best revenge is massive success.}$\rightarrow$\emph{anger, joy, optimism}

By giving high attention to “revenge”, we can infer the emotion \emph{anger}, but to infer emotions \emph{joy} and \emph{optimism} we need to consider the whole sentence in a different perspective. Therefore, a single attention vector will not give us all the emotions associated.

To overcome the above limitation, we propose a new attention architecture, which we call \emph{Pyramid Attention Network}. In the proposed model, attention layer-1 attends to the word embeddings and GRU layer-1 whereas attention layer-2 attends to both of them and GRU layer-2 in addition. This can be generalized to a n-layer stacked GRU or LSTM as well.

The fundamental concept behind this attention mechanism is inspired by \newcite{bahdanau2014neural}, \newcite{yang2016hierarchical} and \newcite{felbo2017using} where the key difference lies in the attention architecture in general. \newcite{felbo2017using} follows a similar approach but only limits to a single attention layer which attends to all the previous layers. \newcite{yang2016hierarchical} used two attention layers but it contrasts from our model since it uses a word attention layer followed by a sentence embedding layer and subsequently a sentence attention layer. Our model is formally defined as follows. 

For an arbitrary vector $U$ we can define attention coefficients ($a_i$) and output $V$ as follows.

Let $u_i \in U$,
\begin{gather}
V = \sum\limits_i a_iu_i \\
\text{where, }e_i = u_iw_a + b \text{ and } a_i = \dfrac{\exp (e_i)}{\sum\limits_t \exp(e_t)}
\end{gather}

For the first attention layer, $V_1 = (H_1, X)$, and for the second attention layer, $V_2 = (H_2, H_1, X)$ where $H_1$ and $H_2$ are outputs of Bi-GRU layer 1 and layer 2 respectively and $X$ is the word embeddings for the input sentence.

\subsection{Classification and Training}
Output of the attention layers 1 ($V_1$) and 2 ($V_2$) are concatenated ($V$ = ($V_1, V_2$)) and passed into a dense layer with a sigmoid activation which outputs a vector of size 11. It has a value between 0 and 1 for each emotion class. If the value is larger than a threshold value, it is classified as positive. We used 0.5 as this threshold.
\begin{align}
\hat{y} = \sigma(W_dV + b)
\end{align}
Weighted binary cross entropy loss function is used with a weight of $w = 2$ for the correctly labeled ones. $y$ represents the ground truth labels and $\hat{y}$ represents the predicted values. $m$ is the number of emotions, which is 11 in this scenario.
\begin{align}
J(\theta) = -\frac{1}{m} \sum\limits_{i=0}^{m}(wy_i\log(\hat{y_i})+(1-y_i)\log(1-\hat{y_i}))
\end{align}

Regularization is essential to ensure that model is not over-fitted in the training phase. Dropout is a regularization method proposed by  \newcite{JMLR:v15:srivastava14a} which  randomly turn-off a percentage of the neurons of a layer in the network. We apply dropout of 0.2 between the attention layer and the dense layer, spatial dropout of 0.4 between word embedding layer and GRU layer for regularization. We added a Gaussian Noise with 0.1 standard deviation to GRU hidden weights to reduce over-fitting. In addition, we apply  L2 regularization penalty to the loss function to reduce large weights. Moreover, we used early stopping \cite{caruana2001overfitting} where training is stopped after the validation loss stops decreasing. 
The model was trained to minimize the weighted binary cross entropy loss using backpropagation. We used Adam optimizer \cite{kingma2014adam} with a batch size of 64 and an initial learning rate of 0.001, which will be reduced by half for every 3 consecutive failures to reduce validation loss with a lower bound of 0.0001. All the values for regularizations were found empirically.

\section{Experiments and Results}
We used the recently published SemEval 2018 Task 1 emotion classification dataset \cite{mohammad2018semeval} for the evaluation of the model. This dataset consists of 10983 tweets which are categorized into 11 emotion categories; \emph{anger, anticipation, disgust, fear, joy, love, optimism, pessimism, sadness, surprise} and \emph{trust}. Each tweet can have multiple emotions, thus multi-label classification mechanisms are employed. As per the SemEval task, the dataset is split into training, development, and testing set with respectively 6838, 886 and 3259 tweets for each set.

We have compared our proposed model with the top four models of the SemEval 2018 Task 1 \cite{baziotis2018ntua,park2018plusemo2vec,kim2018attnconvnet,rozental2018amobee} and our implementation of Hierarchical Attention Network (HAN) approach proposed by \newcite{yang2016hierarchical} adopted to multi-label case. As shown in Table 1, proposed model achieves the state-of-the-art for the emotion detection dataset by \newcite{mohammad2018semeval}.

\begin{table}[]
\centering
\label{res-table}
\begin{tabular}{|p{2.7cm}|ccc|}
\hline \bf Model & \bf Jaccard & \bf Micro & \bf Macro\\
\hline \newcite{baziotis2018ntua}        & 0.579 & - & - \\
\hline \newcite{park2018plusemo2vec}       & 0.576 & 0.692          & 0.497          \\
\hline \newcite{kim2018attnconvnet}        & 0.574 & 0.687          & 0.511          \\
\hline \newcite{rozental2018amobee}        & 0.566 & 0.673          & 0.490          \\
\hline HAN \cite{yang2016hierarchical}  & 0.567 & 0.683          & 0.535          \\
\hline \bf Proposed Model & \bf 0.589 & \bf 0.701          & \bf 0.550 \\
\hline
\end{tabular}
\caption{Comparing results of the proposed model. Proposed model achieves state-of-the-art for the dataset by \newcite{mohammad2018semeval}. \textbf{Jaccard} - mutli-label accuracy (Jaccard accuracy), \textbf{Micro} - Micro-avg F1 score and \textbf{Macro} - Macro-avg F1 score.}
\end{table}

\section{Analysis of Results}
\begin{figure}[h]
    \centering
    \includegraphics[width=0.5\textwidth]{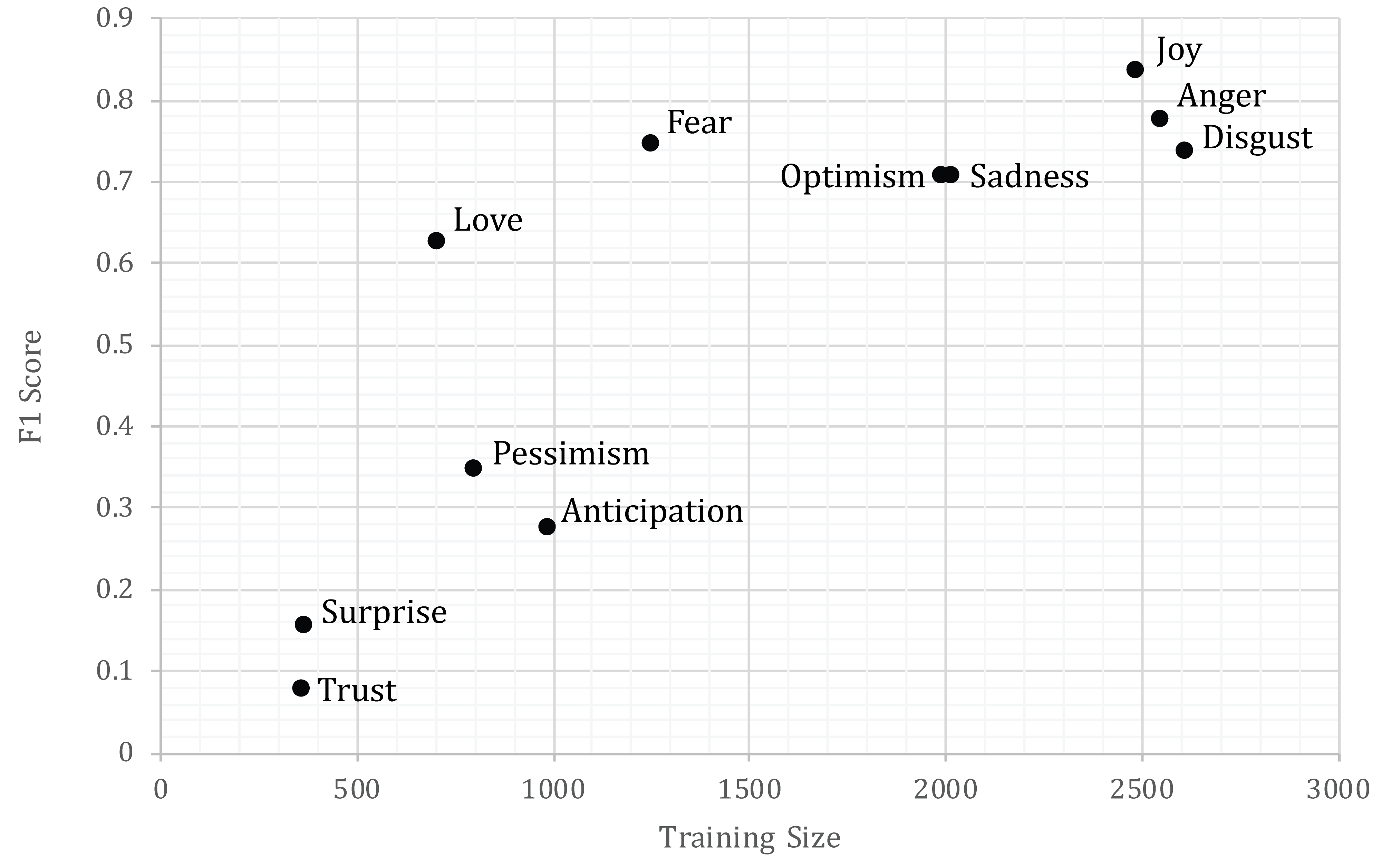}
    \caption{The variation of F1 score with the dataset size of individual emotions}
    \label{fig:chart1}
\end{figure}

Figure \ref{fig:chart1} shows the number of tweets contained in each category vs the measured F-score achieved by our proposed model for each category. Six categories stand out the most with a significantly large amount of data whereas rest of the categories contain a comparably smaller number of tweets. It further shows tweets with a significant number of training examples are detected with a higher F-score whereas underrepresented categories have a lower F-score. Such underrepresented emotions affect the performance of the proposed model severely. For other well represented emotions, our model achieves a F-score more than 0.7. One interesting fact to note is that, though there are comparably smaller number of tweets containing \emph{love}, still our proposed model managed to predict that emotion accurately. The reason for this can be \emph{love} is highly correlated with special keywords.


\section{Conclusion and Future Work}
We proposed a novel \emph{Pyramid Attention Network} (PAN) which achieves state-of-the-art performance for emotion detection in microblogs. Our analysis revealed that there are underrepresented categories in the dataset and those classes affect the model accuracy significantly. Future work includes enhancing the overall performance by improving the detection of underrepresented classes, experiment this approach for multiple emotion detection datasets, and apply this attention mechanism to other text classification tasks.
\bibliography{emnlp2018}
\bibliographystyle{acl_natbib_nourl}

\end{document}